\documentclass[10pt,twocolumn,letterpaper]{article}

\usepackage{isba}
\usepackage{times}
\usepackage{epsfig}
\usepackage{graphicx}
\usepackage{amsmath}
\usepackage{amssymb}
\usepackage{algorithm,algorithmic}
\usepackage{caption}
\usepackage{subcaption}
\usepackage{url}
\usepackage{booktabs}
\usepackage{marvosym}

\isbafinalcopy 

\def\ie{\textit{i.e.,~}}
\def\etal{\textit{et al.}~}
\def\Vec#1{{\boldsymbol{#1}}}
\def\Mat#1{{\boldsymbol{#1}}}

\ifisbafinal\fi
\begin{document}

\title{Determining the best attributes for surveillance video keywords generation}

\author{Liangchen Liu\\
{\tt\small l.liu9@uq.edu.au}
\and
Arnold Wiliem\\
{\tt\small a.wiliem@uq.edu.au}
\and
Shaokang Chen\\
{\tt\small shaokangchenuq@gmail.com}
\and
Kun Zhao\\
{\tt\small k.zhao1@uq.edu.au}
\and
Brian C. Lovell\\
{\tt\small lovell@itee.uq.edu.au}\\
The University of Queensland, School of ITEE\\
QLD 4072, Australia\\
}

\maketitle
\thispagestyle{empty}

\begin{abstract}
	\label{abstract}
	Automatic video keyword generation is one of the key ingredients in reducing the burden of security officers in analyzing surveillance videos.
	Keywords or attributes are generally chosen manually based on expert knowledge of surveillance.
	Most existing works primarily aim at either supervised learning approaches relying on extensive manual labelling or hierarchical probabilistic models that assume the features are extracted using the bag-of-words approach; thus limiting the utilization of the other features. 
	To address this, we turn our attention to automatic attribute discovery approaches.
	However, it is not clear which automatic discovery approach can discover the most meaningful attributes. 
	Furthermore, little research has been done on how to compare and choose the best automatic attribute discovery methods. 
	In this paper, we propose a novel approach, based on the shared structure exhibited amongst meaningful attributes, that enables us to compare between different automatic attribute discovery approaches.
	We then validate our approach by comparing various attribute discovery methods such as PiCoDeS on two attribute datasets.
	The evaluation shows that our approach is able to select the automatic discovery approach that discovers the most meaningful attributes.
	We then employ the best discovery approach to generate keywords for videos recorded from a surveillance system. 
	This work shows it is possible to massively reduce the amount of manual work in generating video keywords without limiting ourselves to a particular video feature descriptor.
\end{abstract}

\section{Introduction}
\label{introduction}

Automatic video analytics is one of the key components in smart surveillance systems to combat crime and terrorism. 
For example, they can be used to detect anomalous events to alert security officers~\cite{Ko2008survey}.
In general, surveillance systems generate a large amount of video data.
This makes finding critical information in surveillance video as challenging as finding the proverbial needle in a haystack~\cite{PritchAVSS09}.
Thus automation is highly desirable so one can reduce the amount of time to find this critical information.

Automatic video analytics have been gaining significant interest in the research community.
Some examples of the current works are: action recognition~\cite{SaninWACV2013}, face hallucination~\cite{liu2012novel}, anomaly detection~\cite{MahadevanCVPR2010}, video description~\cite{meghdadi2013interactive} and video complex event detection~\cite{ChangYYH15,ChangYXY15a}. 

In this work we tackle the problem of automatic generation of keywords for video description.
Keywords are important ingredients in generating textual descriptions~\cite{RohrbachICCV2013}.
More specifically, once the keywords of a video are generated, the video can be searched using natural language to find events of interest. 

\vspace{-.0em}
\begin{figure*}	
	\centering
	\includegraphics[width=0.75\textwidth]{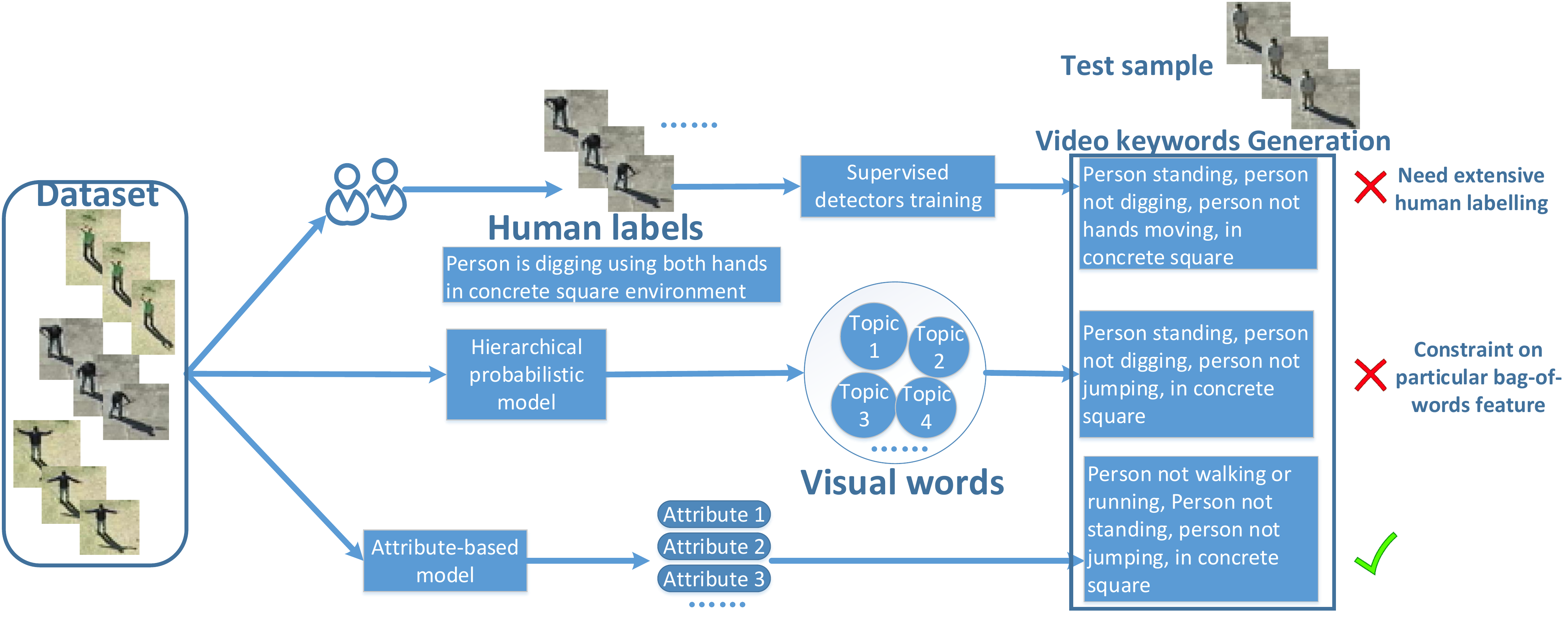}
	\caption{
		The comparisons and properties between our approach and other existing methods on video keywords generation.
	}\label{fig:eyecandy}
	\vspace{-1.4em}
\end{figure*}

Unfortunately, existing approaches still require a great deal of manual labelling before the systems can be used to generate the keywords/description~\cite{hu2007semantic}.
For example, the work proposed by Izadinia~\etal in~\cite{izadinia2012recognizing} uses extensive spatio temporal annotations to train action and role models for action recognition. 
The approach produces better descriptions than many other approaches.
However, the significant manual labelling severely restricts its scalability.
In addition, when relevant manual labels are not available, then it is not possible for the system to describe unusual events which would be extremely useful in anomaly event analysis.

One feasible way to circumvent this is to employ latent hierarchical probabilistic models such as probabilistic Latent Semantic Analysis~(pLSA)~\cite{varadarajan2013sequential} or Latent Dirichlet Allocation~(LDA)~\cite{wang2009unsupervised}.
These methods can automatically mine the latent topics which could represent keywords. Thus, when a topic is inferred in a video, then the associated text of the topic becomes the keyword.
Unfortunately, despite their potential, these methods are based on the bag of words model requiring explicit modelling of visual words. 
Here, each video is assumed to have a collection of visual words. This explicit assumption may not be feasible for other recent video features not derived from bag of word features. 

Inspired from the probabilistic latent topic discovery methods, in this work, we propose a method that can automatically discover video keywords with significantly less manual processing.
More specifically, several attribute discovery methods such as PiCoDeS~\cite{bergamo2011picodes} and Spectral Hashing~\cite{weiss2009spectral} can be employed. 

Visual attribute features are binary features indicating the presence/absence of  visual concepts. 
For instance a car can be described as ['has wheels', 'is metallic', 'does not have legs']. 
In practice, we can represent the binary features as [1 1 0]. 
The attribute features trained in one domain can be reused for another domain with minimum manual work~\cite{Farhadi09describingobjects}. 
As such, a system can be potentially trained to recognize unseen events~\cite{Lampert13}.

Visual attributes have shown promising results in many works which deal with video related tasks~\cite{RohrbachICCV2013,RamanathanICCV2013Video} as well as in some novel problems such as the zero shot learning problem~\cite{Lampert13}.



Once the attribute features are trained, they can be used to extract keywords.
Unfortunately, training attributes features also require extensive manual labeling work. 
This is because, as each individual visual attribute is a binary classifier, then one needs to create the labeled training set for each attribute.

To that end, some researchers have turned their attention to automatic attribute discovery methods~\cite{bergamo2011picodes,rastegari2012attribute,Wiliem2014discovering}.
These methods primarily focus on learning an embedding function that maps the original descriptors into binary code space wherein each individual bit is expected to represent a visual attribute.
We note that these approaches are also closely related to hashing approaches~\cite{gong2011iterative,Leskovec2009Mining,weiss2009spectral}.
The difference is that unlike automatic attribute discovery approaches,
hashing methods are primarily aimed at significantly reducing computational complexity and storage whilst maintaining system accuracy.
Despite many works that have been proposed, it is not clear which methods produce the most meaningful attributes.

Here, we present an approach that allows us to select the attribute discovering method that discovers the most meaningful attributes.
We then find the keywords extracted from the best method to describe videos recorded from a surveillance system.

The intuition of our approach comes from a speculation proposed in~\cite{parikh2011interactive,parikh2011interactively}. More specifically, Parikh et. al. suggest that meaningful attributes tend to occupy a subspace, here called the Meaningful Subspace, on a manifold. 
Thus, we can utilize any given set of meaningful attributes to be our `yardstick' for comparing various attribute discovery methods.

Fig. ~\ref{fig:eyecandy} illustrates the differences between our attribute-based keyword generation approach and the existing approaches.
We can see our approach has two main advantages. 
First, is it does not require significant manual processing. 
Second, it is not constrained to one particular video feature.


\noindent
\textbf{Contributions.} We list our contributions as follows:
(1) We propose an attribute-based video keyword generation approach by utilizing the attribute discovery method that discovers the most meaningful set of attributes; 
(2) To determine the attribute discovery method, we propose a selection approach enabling us to select which attribute discover methods provide meaningful attributes; 
(3) We use and validate our selection method in two known attribute datasets; 
(4) Finally, we validate the keywords extracted from the best attribute discovery method. These keywords can be used to describe videos recorded from a surveillance system.

We continue our paper as follows. 
Section 2 discusses related works.
Section 3 presents our proposed approach to compare various attribute discovery methods. 
Section 4 describes the approach to generate video keywords using discovered attributes. 
Section 5 presents experiments and the section 6 concludes the discussion.

\section{Related Works}
\label{related_works}
There are several methods proposed recently that deal with video keyword and description extraction~\cite{RohrbachPR2014,RohrbachICCV2013,wang2009unsupervised,varadarajan2013sequential,XuXun2015,ChangYLZH16,ChangYHXY15}.
For instance,
Rohrbach~\etal~\cite{RohrbachICCV2013} proposed to generate a rich semantic representation of the visual content such as
object and activity labels.
They employed the Conditional Random Field (CRF) to model all the input visual components. 
In~\cite{RohrbachPR2014}, they extended their work to a three-level-of-detail video description scheme.
Then they applied a machine translation framework to generate the natural language using the semantic representation as sources. 
Unfortunately, this model cannot be used to address our problem due to the extensive manual labelling work required.

To that end, some researchers rely on hierarchical probabilistic models.
Wang~\etal~\cite{wang2009unsupervised} and Varadarajan~\etal~\cite{varadarajan2013sequential} employ LDA and pLSA respectively to perform unsupervised activity analysis.
However as mentioned, these methods can only be applied on the bag-of-words framework.
This means, more powerful features such as Fisher vectors~\cite{perronnin2010large} cannot be used directly.

To the best of our knowledge, there is only one work that specifically targets automatic video description problems in surveillance videos. 
Xu~\etal~\cite{XuXun2015} develop a novel distributed multiple-scene global understanding framework that clusters surveillance scenes by their ability to explain each other’s behaviours.
However, their work only focuses on multiple-scene case and again, utilizes hierarchical probabilistic models.


\section{Selecting the attribute discovery method}
\label{measuring_attribute_meaningfulness}

We first describe the manifold space where the attributes lie. 
Then, we use this representation to select the attribute discovery method that discovers the most meaningful attributes.
Technically, we will measure the meaningfulness of a set of discovered attributes.

\subsection{The manifold of decision boundaries}
\label{sec:manifold}

A visual attribute, or simply an attribute, can be represented as a decision boundary, 
as it partitions a given set of N images $\mathcal{X} = \{ \Vec{x}_i \}_{i=1}^N$ into two subsets, $\mathcal{X}^+ \cup \mathcal{X}^- = \mathcal{X}$: 
(1)~the set of images/videos where the attribute exists, $\mathcal{X}^{+}$; and 
(2)~the set of images/videos where the attribute does not exist, $\mathcal{X}^{-}$.
Hence, in this case, we assume that all attributes lie on a manifold of decision boundaries~\cite{parikh2011interactively}.



In our work, we represent an attribute as an N-dimensional binary vector 
whose $i$-th element is the outcome of the corresponding attribute binary classifier tested on image $\Vec{x}_i$.
Let us consider the corresponding attribute classifier $\phi ( \cdot ) \in \mathbb{R}$,
the function $\phi$ classifies an input image $\Vec{x}_i$ into either the positive or negative set (\ie $\mathcal{X}^+$ or $\mathcal{X}^-$), depending on the sign of the classifier output.
We define $\Vec{z}^{[\mathcal{X}]}$ as
the attribute representation w.r.t. a set of images $\mathcal{X}$, $\Vec{z}^{[\mathcal{X}]} \in \{-1,+1\}^{N}$, 
where the $i$-th dimension of $\Vec{z}_i^{[\mathcal{X}]} = \operatorname{sign}(\phi(\Vec{x}_i)) \in \{-1,+1\}$.
For the sake of clarity, we write $\Vec{z}^{[\mathcal{X}]}$ as $\Vec{z}$, whenever the context is clear.

Thus, the manifold of decision boundaries w.r.t. $\mathcal{X}$ is 
then defined as the lower dimensional space embedded in an N-dimensional 
binary space, $\mathcal{M}^{[\mathcal{X}]} \in \{-1,+1\}^{N}$.
Again, we write $\mathcal{M}^{[\mathcal{X}]}$ as $\mathcal{M}$
whenever the context is clear.

\subsection{Distance from the Meaningful Subspace}
\label{sec:distance}

Given a set of images $\mathcal{X}$ and a set of discovered attributes, $\mathcal{D} = \{\Vec{z}_k\}_{k=1}^K, \Vec{z}_k \in \{-1,+1\}^N$, 
here our goal is to 
define the distance of the attribute set from the Meaningful Subspace.
Ideally, this subspace contains all possible meaningful attributes.
Unfortunately, it is impossible to enumerate all of them.
One possible solution is to use previously human labelled attributes in various image datasets such as~\cite{biswas2013simultaneous,parikh2011interactive,parikh2011interactively}.
These attributes are considered to be meaningful as human annotators labelled them via the Amazon Mechanical Turk~(AMT).
We define this set of meaningful attribute as  $\mathcal{S} = \{\Vec{h}_j\}_{j=1}^J, \Vec{h}_j \in \{-1,+1\}^N$.


Since meaningful attributes are assumed to have \textit{shared structure}, 
we could assume that a meaningful attribute must be able to be 
described using the other meaningful attributes.
For instance, a set of attributes of primary colors red, green and blue could be used
to reconstruct the set of secondary colors such as yellow, magenta and cyan.
The primary colors could also be used to reconstruct the other primary colors (\eg red is not green and not blue).

Unfortunately, a linear combination of
attributes may not lie on the manifold $\mathcal{M}$.
More precisely, $\sum_i w_i \Vec{z}_i$ may not be a member of $\mathcal{M}$ as it is possible that 
$\sum_i w_i \Vec{z}_i \notin \{-1,+1\}^N$, rather $\sum_i w_i \Vec{z}_i \in \mathbb{R}^N$.
Thus, it is non-trivial to calculate the geodesic distance (\ie the shortest distance between two points on the manifold) for determining the distance between a discovered attribute, $\Vec{z}_k$ and the Meaningful Subspace, $\mathcal{S}$.
Therefore, we consider an approximated geodesic distance by assuming the members of manifold $\mathcal{M}$ lie in $\mathbb{R}^N$. 
In this case both,
magnitude and sign of the classifier output values are considered.
Thus, the approximated geodesic distance is defined as:
\begin{equation}
\min_\Vec{r} \| \Mat{A} \Vec{r}  - \Vec{z}_k \|_2^2,
\label{eq:single_dist}
\end{equation}
\noindent
where the matrix $\Mat{A} \in \mathbb{R}^{N \times J}$ contains the attributes of set $\mathcal{S}$ arranged as column vectors; 
$\Vec{r} \in \mathbb{R}^{J \times 1}$ is the reconstruction coefficient vector.
The above distance is defined in terms of the reconstruction error of the attribute $\Vec{z}_k$ by the set of meaningful attributes $\mathcal{S}$.
When an attribute is meaningful, then its reconstruction error is minimized or close to zero 
due to the \textit{shared structure} possessed by the Meaningful Subspace.

We then define the distance between the set of discovered attributes $\mathcal{D}$ and the Meaningful Subspace $\mathcal{S}$ on the manifold $\mathcal{M}$ w.r.t. the set of images $\mathcal{X}$ as the average reconstruction error:
\begin{equation}
\delta(\mathcal{D},{\mathcal{S}}; \mathcal{X}) = \frac{1}{K}\min_{\Mat{R}} \| \Mat{A}\Mat{R} - \Mat{B} \|_F^2,
\label{eq:min_dist}
\end{equation}
where $\| \cdot \|_F$ is the matrix Frobenious norm;
$\Mat{B} \in \mathbb{R}^{N \times K}$ is the matrix of attributes $\mathcal{D}$ arranged as column vectors;
$\Mat{R} \in \mathbb{R}^{J \times K}$ is the reconstruction matrix.

The distance in~\eqref{eq:single_dist} and~\eqref{eq:min_dist} may create dense 
reconstruction coefficients, suggesting that each meaningful attribute 
should contribute to the reconstruction.
A more desired result is to have less dense coefficients (\ie less number of non-zero coefficients).
This is because there may be only a few meaningful attributes required 
to reconstruct another meaningful attribute.
One possible way to address this is to add 
the convex hull regularization which has been shown in~\cite{cevikalp2010face} to induce sparsity.

When a convex hull constraint is considered,~\eqref{eq:min_dist} becomes:
\begin{equation}
\label{eq:con_hull}
\begin{aligned}
\delta_{\operatorname{cvx}}(\mathcal{D},\mathcal{S}; \mathcal{X}) = \frac{1}{K}\min_{\Mat{R}} \| \Mat{A}\Mat{R} - \Mat{B} \|_F^2   \operatorname{s.t.} \\
{\Mat{R}}\left( {i,j} \right) \ge 0 & \\
\sum_{i=1}^{J} {\Mat{R}(i,\cdot)}=1. &
\end{aligned}	
\end{equation}
\noindent
The above equation basically computes the average distance between each discovered attribute $\Vec{z}_k \in \mathcal{D}$ and the convex hull of $\mathcal{S}$.
The above optimization problem could be solved using the method proposed in~\cite{cevikalp2010face}.
In our approach, we assume that the lower the distance of a set of discovered attributes to a meaningful subspace, the more meaningful the attributes will be.


\section{Generating keywords using discovered attributes}
\label{surveillance_video_description}
Once meaningful attributes are discovered, one can extract the attribute features from the given data.
However, one still needs to name the attributes. 
Despite this manual process, we argue that the manual process for naming meaningful attributes is significantly easier and quicker than the manual process of labelling images/videos to train attribute features.

One can name an attribute by first extracting the attribute features from a given set of images.
As previously mentioned, each attribute divides any set of images/videos into two groups: the group of images in which the visual attribute is present (the positive class)
and the group of images/videos in which the visual attribute is absent (the negative class).

Some attributes may have similar names. In this case, these attributes are considered as duplicate and therefore they are merged.

\section{Experiment}
\label{experiment}

In this section, we validate our proposed approach and evaluate the accuracy of the keywords extracted from the best discovered method to describe videos.

In the first part, we evaluate the ability of our approach to measure the meaningfulness of a set of attributes. 
Then, we use our proposed approach to evaluate attribute meaningfulness on
the
attribute sets generated from various automatic attribute discovery methods such as
PiCoDeS~\cite{bergamo2011picodes} as well as
the hashing methods such as
Spectral Hashing (SPH)~\cite{weiss2009spectral} and
Locality Sensitivity Hashing (LSH)~\cite{Leskovec2009Mining}.
For this case, two datasets will be utilized: (1) a-Pascal a-Yahoo dataset
(ApAy)~\cite{Farhadi09describingobjects}; (2) SUN Attribute
dataset (ASUN)~\cite{Patterson2012SunAttributes}.

In the second part of our experiment,
we apply the best attribute discovery method to discover keywords from a surveillance dataset. In this setting, we utilize the UT Tower aerial view dataset (UTTower)~\cite{UT-Tower-Data}. 
The efficacy of the keywords are then evaluated.

\subsection{Datasets and experiment setup}
The following are the detailed description of each image dataset for validating our approach and evaluating the attribute discovery methods.

\textbf{a-Pascal a-Yahoo dataset (ApAy)~\cite{Farhadi09describingobjects} --- } comprises two sources: a-Pascal and a-Yahoo.
There are 12,695 cropped images in a-Pascal that are divided into
6,340 for training and 6,355 for testing with 20 categories.
The a-Yahoo set has 12 categories disjoint from the a-Pascal categories.
Moreover, it only has 2,644 test exemplars.
There are 64 attributes provided for each cropped image.
In total the dataset has 15,339 exemplars, 64 attributes and 32 categories.
The dataset provides four features for each exemplar: local texture; HOG; edge and color descriptor.
These are then concatenated into a 9,751 dimensional feature vector.
We use the training set for discovering  attributes and we perform our study in the test set. More precisely, we consider the test set as the set of images $\mathcal{X}$.

\noindent
\textbf{SUN Attribute dataset (ASUN)~\cite{Patterson2012SunAttributes} --- } ASUN is a fine-grained scene classification dataset
consisting of 717 categories (20 images per category) and 14,340 images in total with 102 attributes.
There are four types  of features provided in this dataset: (1) GIST; (2) HOG;
(3) self-similarity and (4) geometric context color histograms~(See \cite{xiao2010sun} for feature and kernel details).
From 717 categories, we randomly select 144 categories for discovering attributes.
As for our evaluation, we random select 1,434 images (\ie 10\% of 14,340 images) from the dataset.
It means, in our evaluation, some images may or may not come from the 144 categories used for discovering attributes.

For the first experiment, we apply the following pre-processing described in~\cite{bergamo2011picodes}.
We first lift each feature into a higher-dimensional
space approximating the histogram intersection kernel by using the
explicit feature maps proposed by Vedaldi and Zisserman~\cite{Vedaldi2012Efficient}.
More precisely, each feature is mapped into the space three times larger than the original
space.
This effectively allows us to apply linear classifiers in the explicit kernel space~\cite{bergamo2011picodes}.
After the features are lifted, we then apply PCA to reduce the dimensionality of the feature space by 40 percent.
This pre-processing step is crucial for PiCoDeS as it uses lifted feature space to simplify their training scheme while maintaining the information preserved in the Reproducing Kernel Hilbert Space~(RKHS).
Therefore, the method performance will be severely affected when lifting features are not used.
In our empirical observations (results not presented), we also found that lifted feature space gives positive contributions to the other methods.

Each method is trained using the training images to discover the attributes.
Then we use the manifold $\mathcal{M}$ w.r.t. the test images for the evaluation.
More precisely, each attribute descriptor is extracted from test images (\ie $\Vec{z}_k, \Vec{z}_k \in \{-1,1\}^N$, where N is the number of test images).
For each dataset, we use the attribute labels from Amazon Mechanical Turk~(AMT) to represent the Meaningful Subspace, $\mathcal{S}$.

\textbf{UT Tower aerial view activity classification dataset (UTTower)~\cite{UT-Tower-Data} --- } consists of 108 low-resolution video sequences from 9 types of actions. Each action is performed 12 times by 6 individuals. The dataset is composed of two types of scenes: concrete square and lawn.  There are 4 actions in the concrete square scene, they are ``pointing'', ``standing'', ``digging'', ``walking'' and 5 actions in the lawn scene: ``carrying'', ``running'', ``wave1'', ``wave2'', ``jumping''.
Ground truth labels for all actions videos are provided for the training and the testing.

For the second experiment, we use manifold feature proposed in \cite{Zhao2015Kernelised} to extract visual information from the surveillance videos in the dataset.
The video frames were first downsized into 16 $\times$ 16 and then Grassmann points on $\mathcal{G}_{128,8}$ were generated by performing the SVD on the normalized pixel intensities of 8 successive frames. In total, there are 216 manifold points.
Note that, the features are not derived from the bag-of-words framework.
It is also noteworthy to mention that our work is not primarily aimed to study feature discriminative power and robustness. 
Although, it is generally assumed that better features may provide more meaningful attributes, further studies are required in the future.

\subsection{Attribute meaningfulness evaluation}
In this experiment, our aim is to verify whether the proposed
approach does measure meaningfulness on the set of discovered
attributes. 
One of the key assumptions in our proposal is that the meaningfulness is reflected in the distance between the meaningful subspace and the given attribute set, $\mathcal{D}$.
That is, if the
distance is far, then it is assumed that the attribute set is less
meaningful, and vice versa.
In order to evaluate this assumption we create two sets of attributes,
meaningful and non-meaningful attributes, and observe their
distances to the meaningful subspace.

For the meaningful attribute set, we use the attributes from AMT
provided in each dataset. More precisely, given manually labelled attribute set $\mathcal{S}$, we divide the set into two subsets $\mathcal{S}^1
\cup \mathcal{S}^2 = \mathcal{S}$. Following the method used in
Section~\ref{measuring_attribute_meaningfulness}, we use $\mathcal{S}^1$ to represent the
Meaningful Subspace and consider $\mathcal{S}^2$ as a set of
discovered attributes (\ie $\mathcal{D} = \mathcal{S}^2$). As
human annotators are used to discover $\mathcal{S}^2$, these attributes are
considered to be meaningful. We name this as the
\textit{MeaningfulAttributeSet}.

For the latter, we create attributes that are not meaningful by
random generation. 
Note that random generation is important to ensure the division is not subjective. 
More precisely, we generate a finite set of random
attributes $\tilde{\mathcal{N}}$. As the set $\tilde{\mathcal{N}}$ is
non-meaningful, it should have significantly large distance to the Meaningful
Subspace. We name this set as \textit{NonMeaningfulAttributeSet}.
Furthermore, we progressively add random attributes to the set of
attributes discovered from each method, to evaluate whether the distance to Meaningful Subspace is
enlarged when the number of non-meaningful attributes increases.

Fig.~\ref{fig:noise} presents the evaluation results where the methods are configured to discover 32 attributes. 
From the results, it
is clear that \textit{MeaningfulAttributeSet} has the closest
distance to the Meaningful Subspace in all datasets. As expected the
\textit{NonMeaningfulAttributeSet} has the largest distance
compared with the others. In addition, as more random attributes
are added, the distance between the sets of attributes discovered
for every approach and the Meaningful Subspace increases. These
results indicate that the proposed approach could measure the set
of attribute meaningfulness. In addition, these also give a strong
indication that meaningful attributes have the
\textit{shared structure}.

The results presented in Fig.~\ref{fig:noise} suggest that PiCoDeS consistently discovers the most meaningful attributes on both datasets. 
SH is the second best method to discover meaningful attributes.
PiCoDeS utilizes max-margin framework to discover the attributes whereas SH uses spectral relaxation to preserve the similarity between data points in the binary space.
In addition, as expected LSH employing random projection approach, is one of the worst performing methods.


\subsection{Generating video keywords using discovered attributes}
In this experiment, we will follow the strategy proposed in section~\ref{surveillance_video_description}.
Here we ask experts to perform the attribute naming task for the three attribute discovery methods such as PiCoDeS, SH and LSH
configured to discover 16 attributes on the UTTower surveillance video dataset.
Then we will use the named attributes as the keywords.
To make our work reproducible, our experiment results will be available online\footnote{\url{http://www.itee.uq.edu.au/sas/datasets}} after this work is published.
%

Note that we only take into account the attributes that can be named by experts.
This means, any attribute that cannot be named will not be considered as a valid keyword.
After performing this task, we found that there are 9 attributes for PiCoDeS, 8 attributes for SH and 3 attributes for LSH that can be named.
These results suggest that our proposed approach is capable of guiding us in selecting the best attribute discovery methods as the experts are able to name most of the discovered attributes by PiCoDeS and SH.

Once attributes are named, the next step is to generate keywords of each video.
Technically, the attributes are extracted from each video. 
Then, the keywords are generated using the terms of the associated positive attributes..

We evaluate the quality of the generated keywords to describe each video. 
We then ask human experts to determine whether a keyword is suitable to describe a video.

\begin{figure}	
	\centering
	\begin{subfigure}[t]{0.22\textwidth}
		\centering
		\includegraphics[width=1\textwidth]{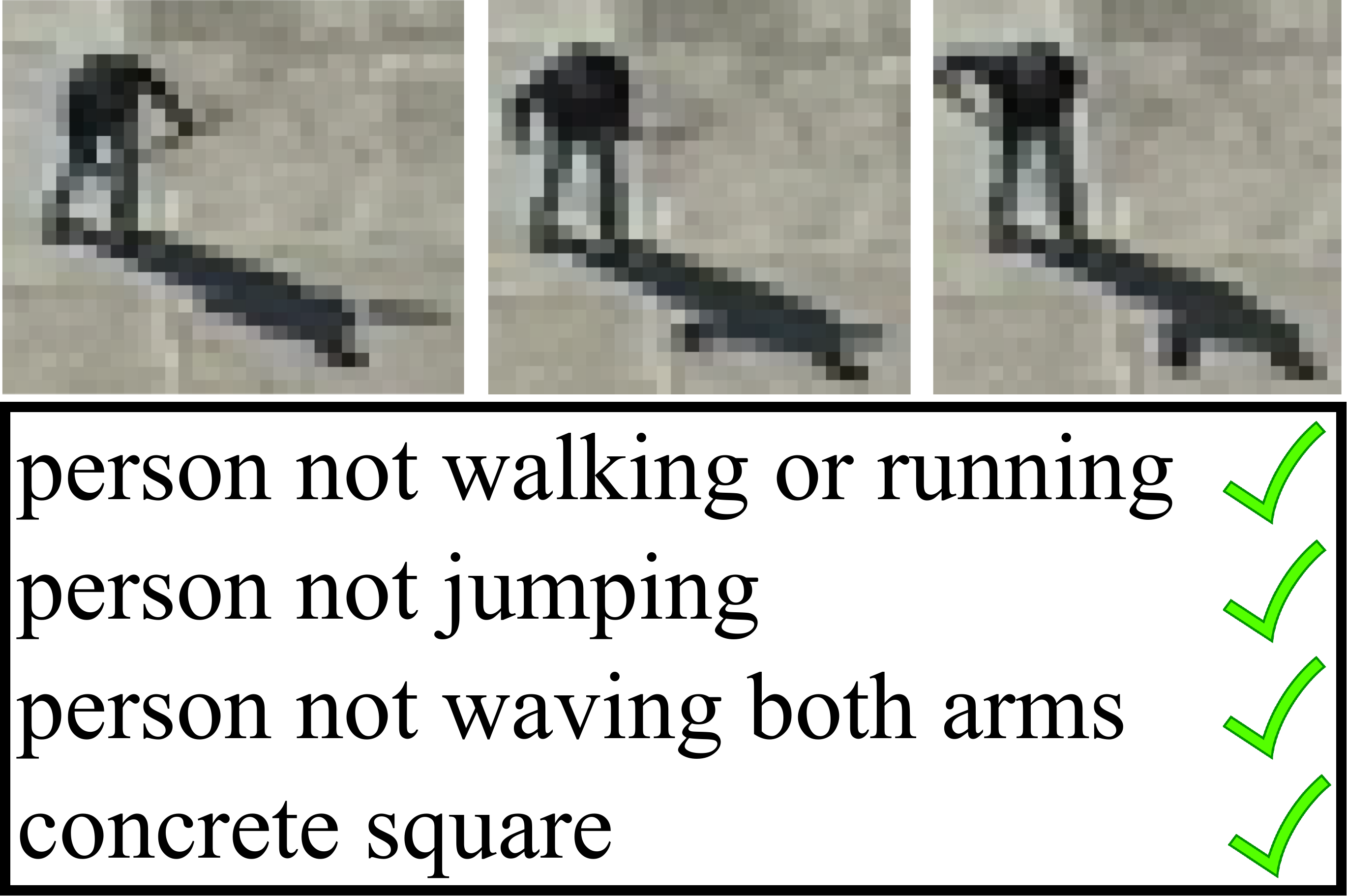}
		\caption{}\label{fig:video1}		
	\end{subfigure}
	\quad
	\begin{subfigure}[t]{0.22\textwidth}
		\centering
		\includegraphics[width=1\textwidth]{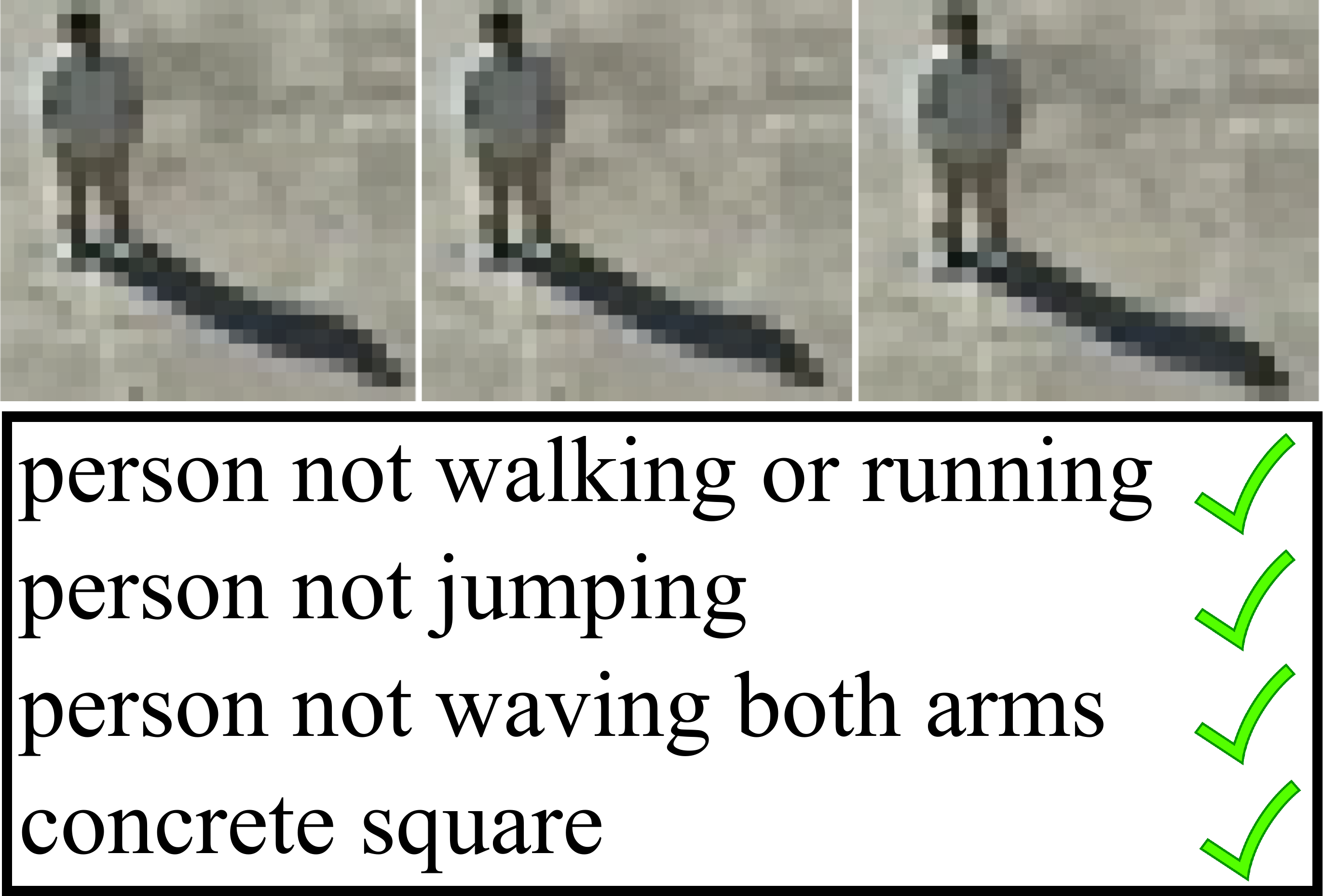}
		\caption{}\label{fig:video2}
	\end{subfigure}
	\quad
	\begin{subfigure}[t]{0.22\textwidth}
		\centering
		\includegraphics[width=1\textwidth]{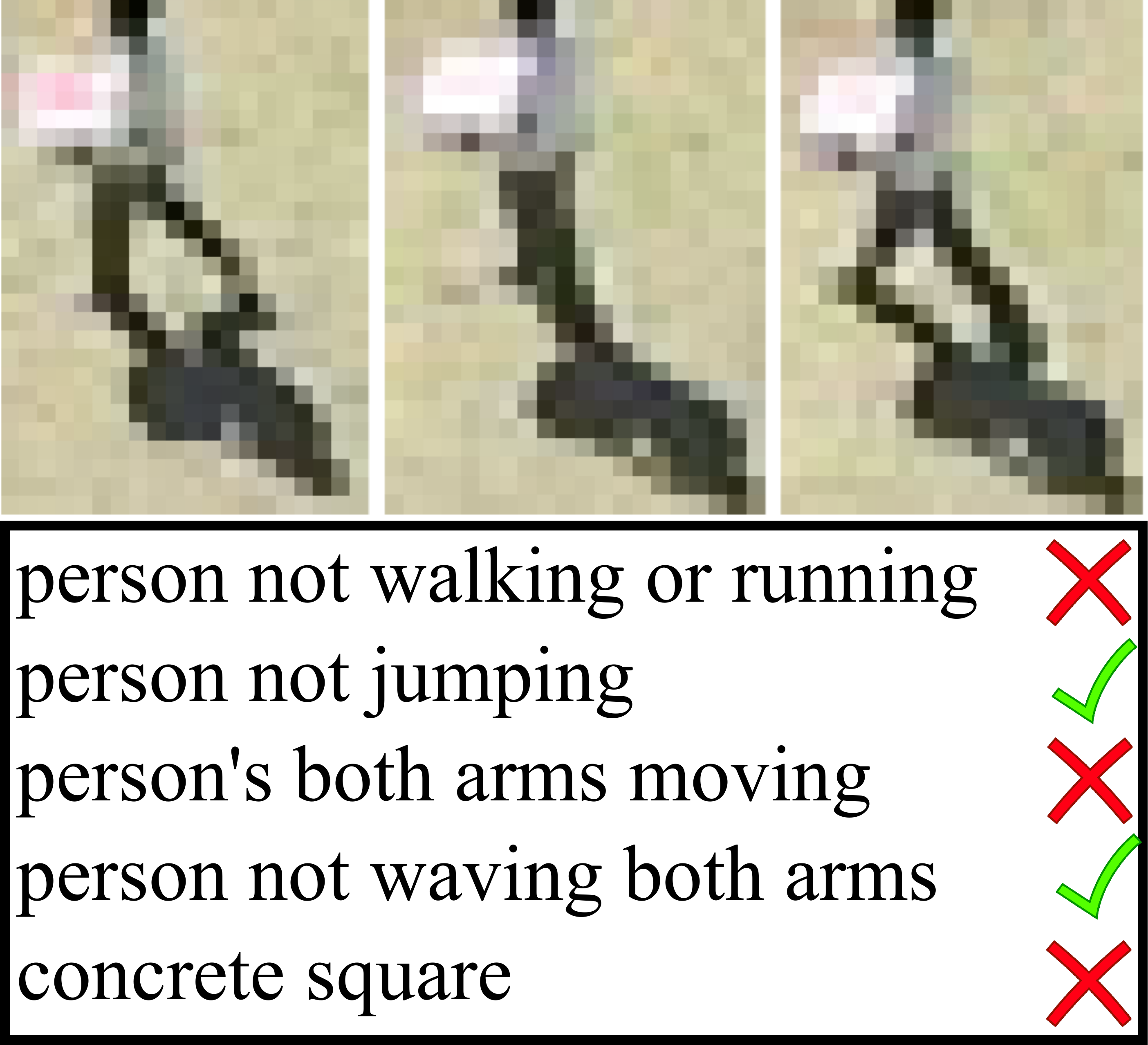}
		\caption{}\label{fig:video3}
	\end{subfigure}
	\quad
	\begin{subfigure}[t]{0.22\textwidth}
		\centering
		\includegraphics[width=1\textwidth]{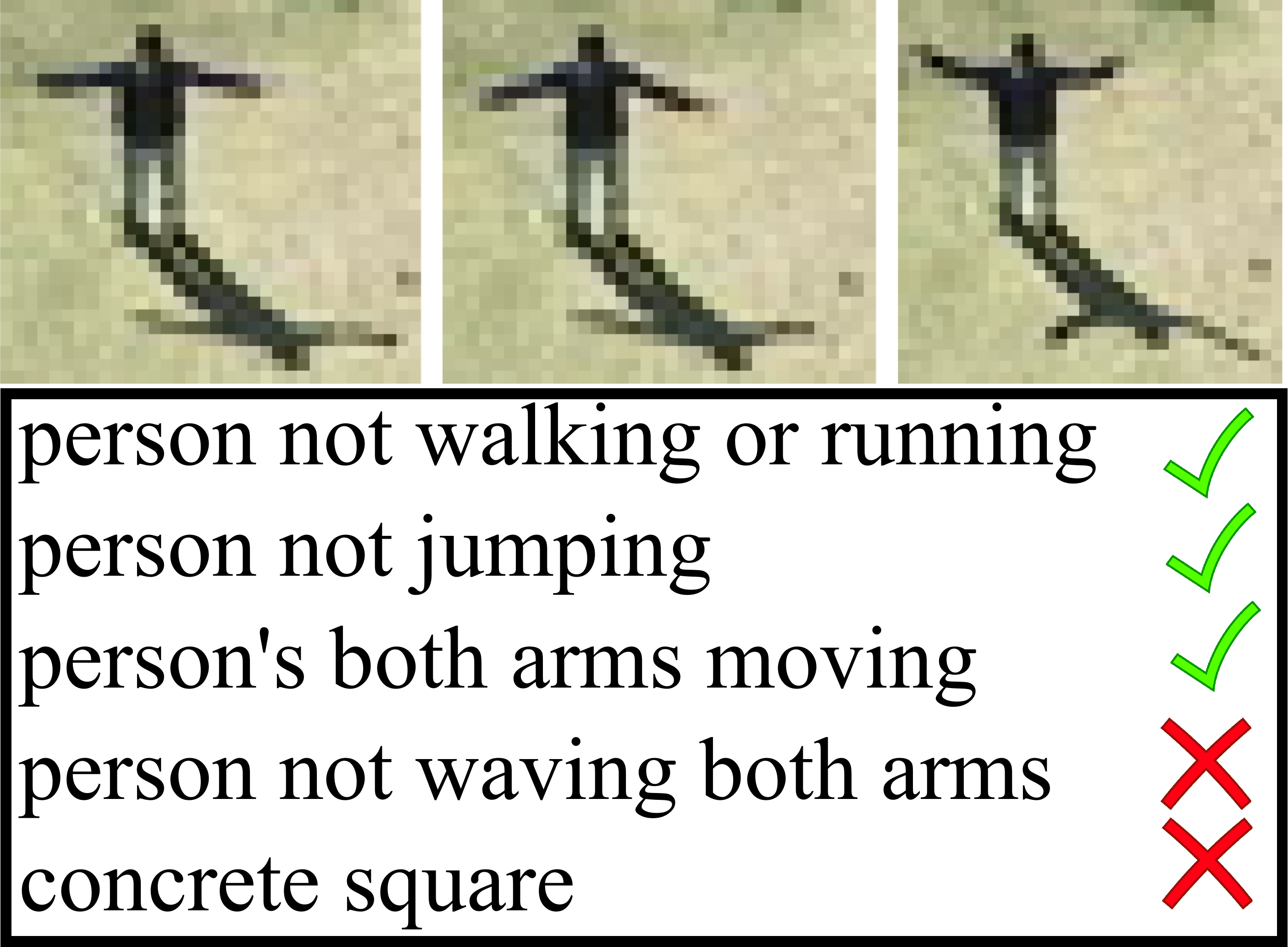}
		\caption{}\label{fig:video4}
	\end{subfigure}
	
	\caption{
		The demonstrations of video description based on attributes from PiCoDeS. 
		(a) and (b) are two samples of videos in which most keywords are suitable; (c) and (d) are two of the worst ones.
	}\label{fig:video}
\end{figure}

Fig.~\ref{fig:video} presents two examples where videos are described with suitable keywords and two examples where videos are described with unsuitable keywords.
The examples depicted in Fig~\ref{fig:video}, (a), (b), (c) and (d) are videos of digging, standing, carrying and waving, respectively.


\begin{figure}	
	\centering
	\begin{subfigure}[t]{0.22\textwidth}
		\centering
		\includegraphics[width=1\textwidth]{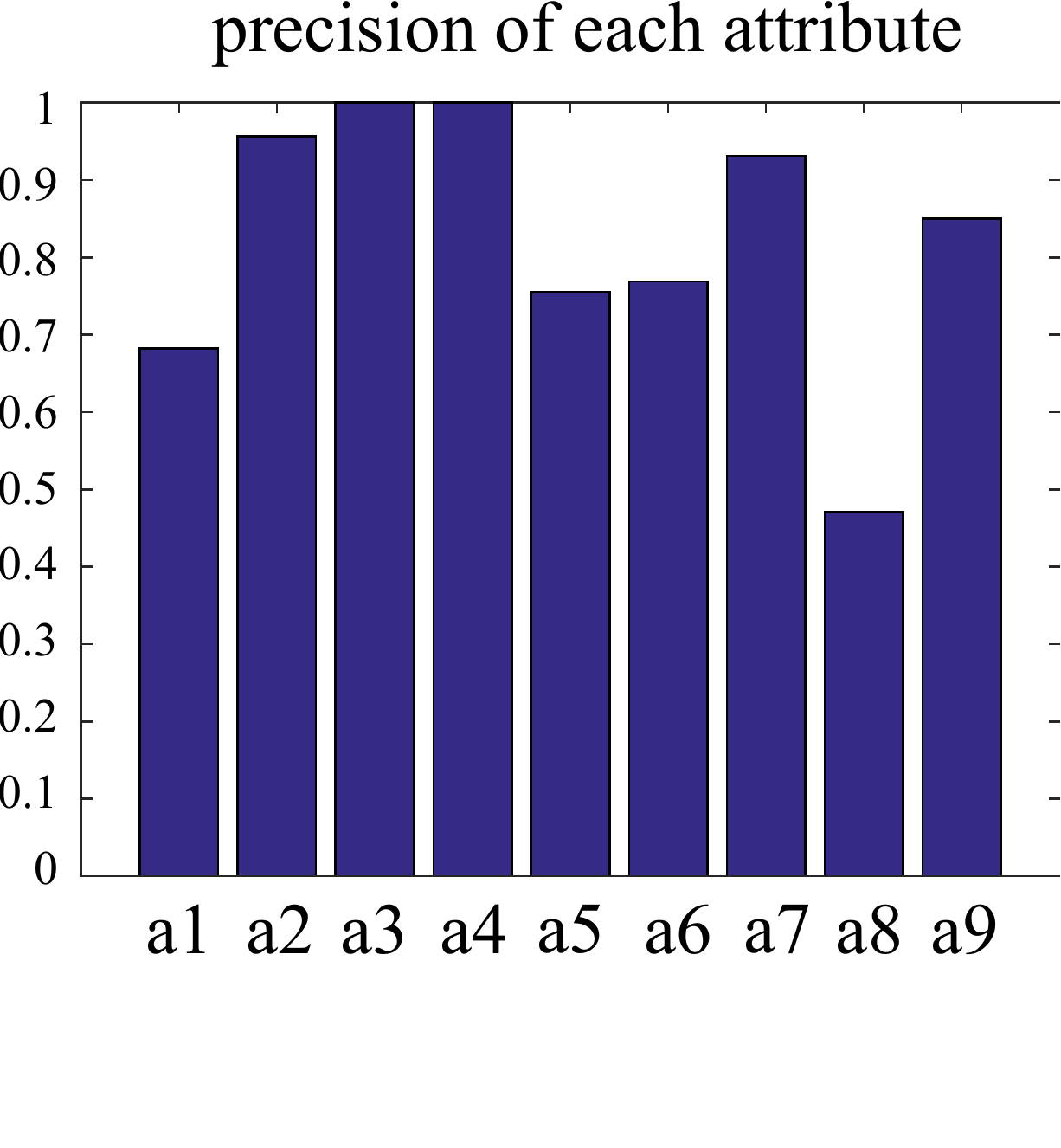}
		\caption{}\label{fig:1a}		
	\end{subfigure}
	\quad
	\begin{subfigure}[t]{0.22\textwidth}
		\centering
		\includegraphics[width=1\textwidth]{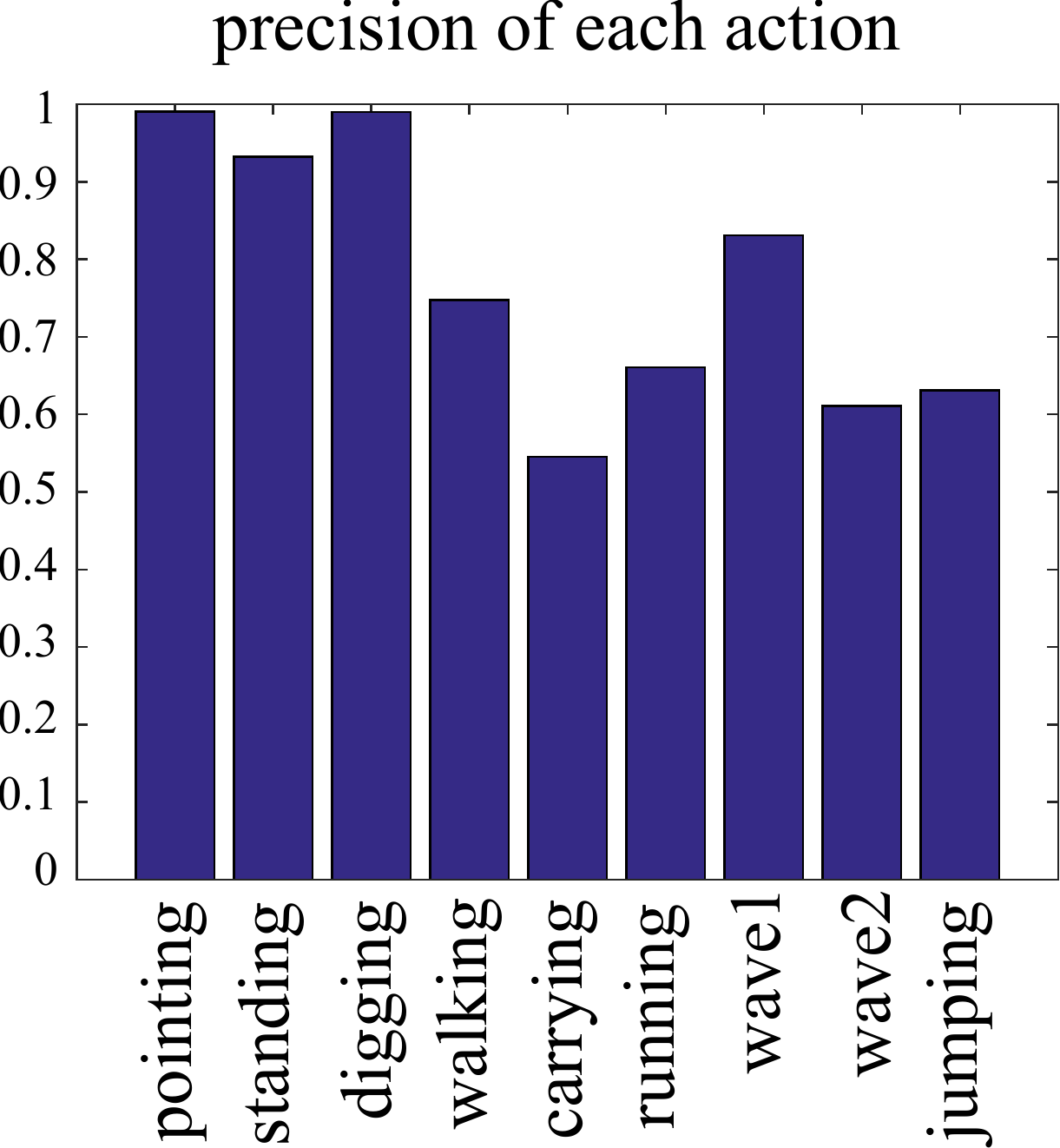}
		\caption{}\label{fig:1b}
	\end{subfigure}
	\quad
	\begin{subfigure}[t]{0.22\textwidth}
		\centering
		\includegraphics[width=1\textwidth]{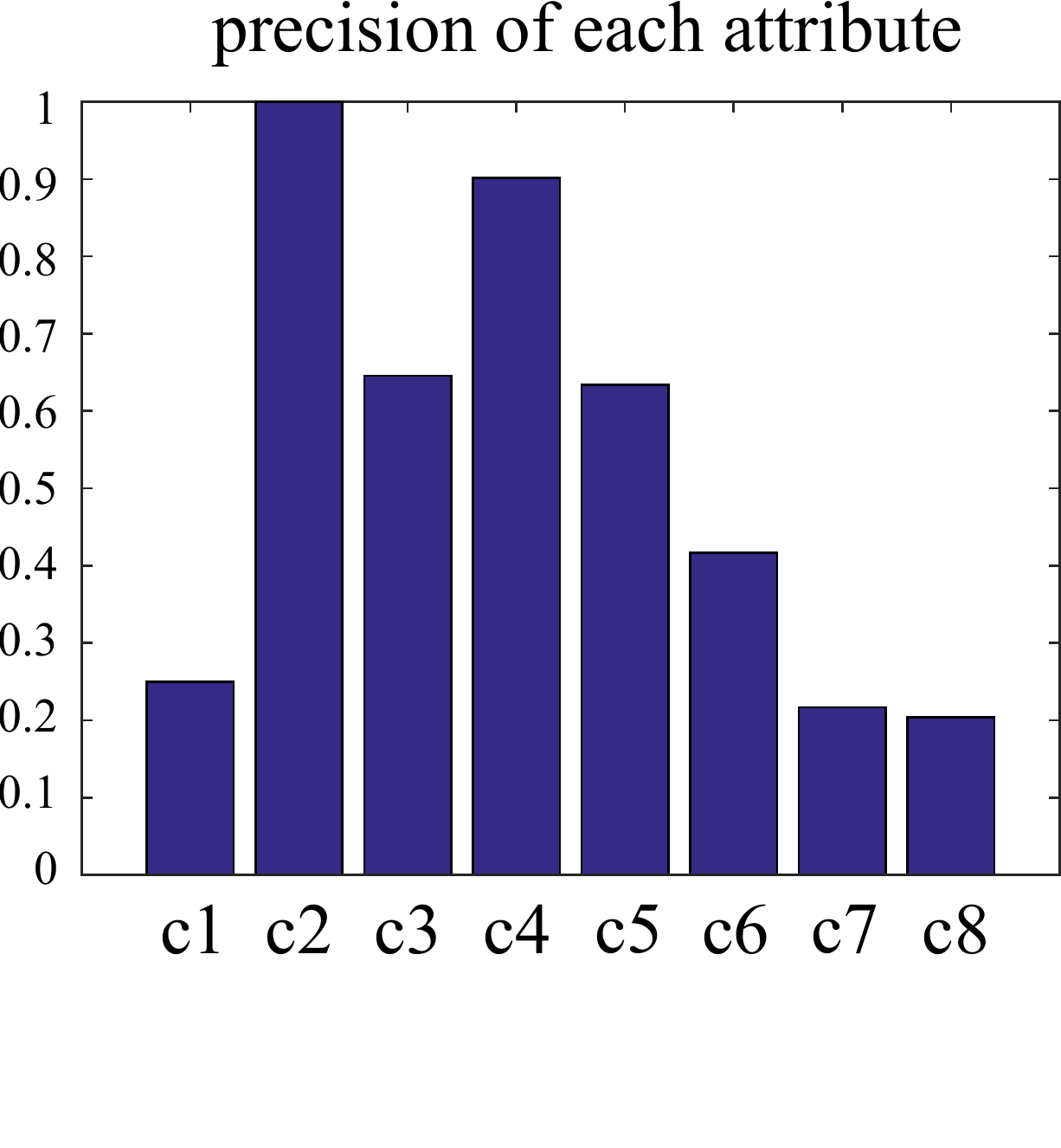}
		\caption{}\label{fig:1c}
	\end{subfigure}
	\quad
	\begin{subfigure}[t]{0.22\textwidth}
		\centering
		\includegraphics[width=1\textwidth]{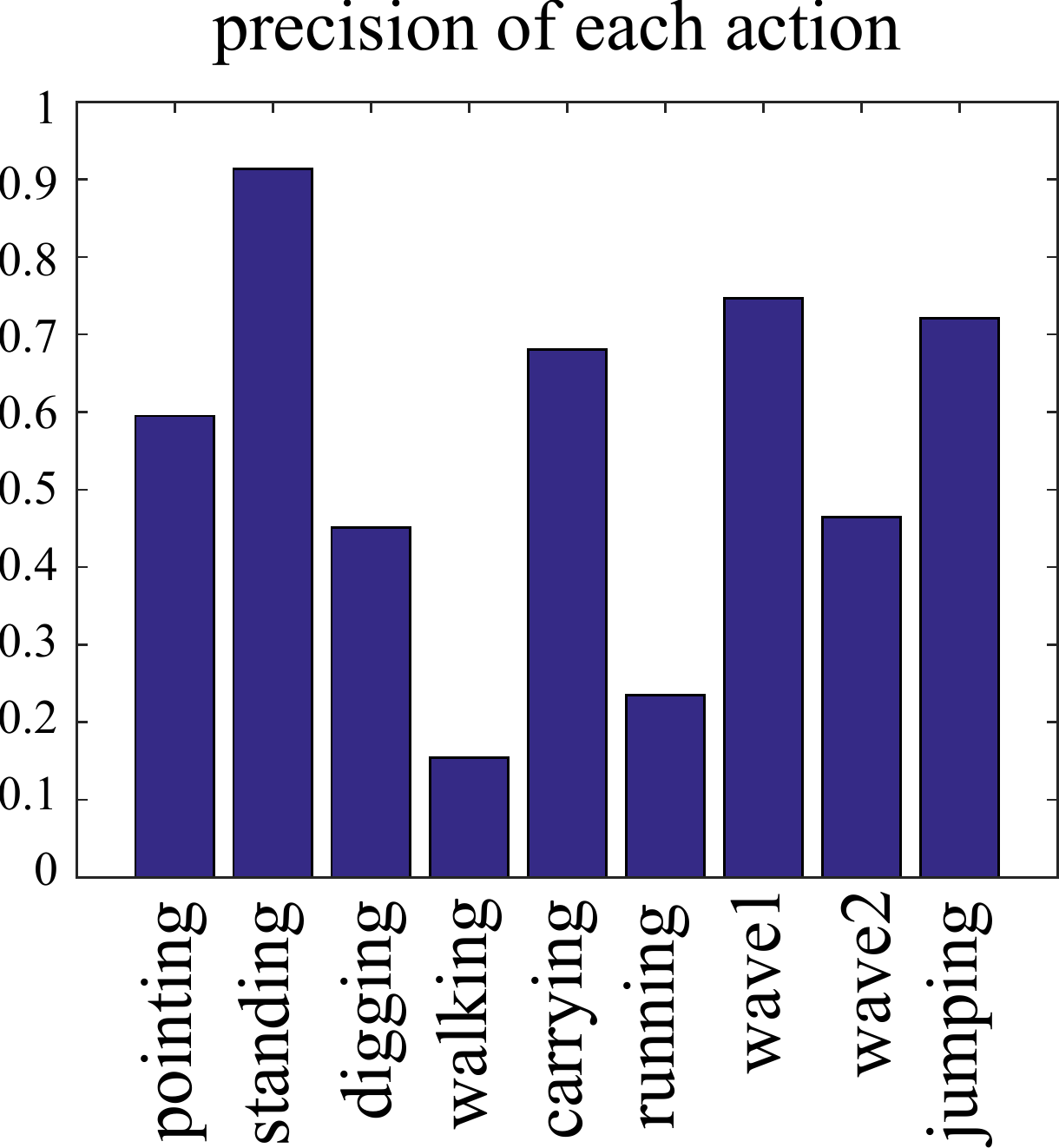}
		\caption{}\label{fig:1d}
	\end{subfigure}
	
	\caption{
		The detailed results of precision of each attribute and precision of each action for PiCoDeS in the first row and
		SH in the second row. 
		The horizontal axis in (a),(c) indicates the ID of discovered attributes and the ones in (b),(d) 
		indicate the ID of actions. 
		The vertical axes represent hit rate~(precision); 
		a1 is 'person not walking or running',
		a2 is 'person not jumping',
		a3 is 'person lower part stationary',
		a4 is 'person's four limbs not moving',
		a5 is 'person's both arms moving',
		a6 is 'person pointing',
		a7 is 'person not waving both arms',
		a8 is 'concrete square',
		a9 is 'lawn scenes',
		c1 is 'person standing',
		c2 is 'person not walking or running',
		c3 is 'person not moving both arms',
		c4 is 'person not carrying',
		c5 is 'person arms not moving separately',
		c6 is 'person not moving arms',
		c7 is 'person holds arm in air',
		c8 is 'person carrying'.
	}\label{fig:precision1}
\vspace{-1.0em}
\end{figure}

We count the number of keywords correctly used in each video description and compute the correct hit rate for the whole testing set. 
The correct hit rate for PiCoDeS, SPH and LSH are 77.7\%, 55.9\% and 48.3\%, respectively. This further validates our proposed approach to measure attribute meaningfulness. In addition, it also shows that using the best attribute discovery method, we can automatically generate keywords for videos in a more economical way.
Fig.~\ref{fig:precision1} presents further results in this evaluation. In particular, (a) and (b) report the hit rate for PiCoDeS of each attribute and action, respectively. The plots in (c) and (d) are the hit rate for SH of each attribute and action, respectively.
Most attributes discovered by PiCoDeS have more than 70\% hit rate with two attributes having 100\% hit rate (all correct). The hit rate for each action also demonstrates an overall good hit rates with most videos being described with hit rate more than 60\%.
The results for SH are worse than PiCoDeS.


\noindent
\textbf{Analysis on cost and time saving in the manual process} Here we compare the time and cost required to perform manual work between our method and the traditional approaches requiring extensive manual processing.
The time and cost analysis is based on the AMT Human Intelligent Task (HIT). 
One HIT normally comprises a set of tasks that human could do to label one image/video data. 
Let $J$ be the number of keywords and $N$ be the number of training samples which is usually very large number. 
In our method, we are only required to name the discovered attributes. 
Hence, our method require just $J$ HITs. 
On the other hand, traditional approaches require at least $N$ HITs as these require all training samples to have the keywords. 
Indeed as $J << N$, then our method massively reduces the time and cost required as it has much less number of HITs.

\section{Conclusion}
\label{conclusion}

In this paper, we described an attribute-based video keyword generation approach. 
Our approach utilized an existing automatic attribute discovery approach to discover the keywords. 
Since there have been numerous attribute discovery approaches in the literature, we devise a selection method, based on the shared structure exhibited amongst meaningful attributes, that enables us to compare the efficacy between different automatic attribute discovery approaches.
In particular, we devised a distance function that measures the meaningfulness of a set of discovered attributes. 
We used our approach to select the methods that are most likely to discover meaningful attributes. 
Then, we validated our approach on two attribute datasets.
The results showed that our approach is able to determine which automatic attribute discovery method can generate the most meaningful keywords or attributes.
Finally, we showed how the discovered attributes were used to generate keywords for videos recorded from a surveillance system.

The proposed approach indicates that it is possible to dramatically reduce the amount of manual work in generating video keywords without limiting ourselves to arbitrary preselected video feature descriptors. 

We note that our proposed selection method only indicates the best attribute discovery method. 
Thus, a more quantitative approach may be required in future study. 
In addition, various regularizations such as the $\ell_1$ regularization for (\ref{eq:single_dist}) and (\ref{eq:min_dist}) will be explored in the future. 
The $\ell_1$ constraint is an explicit regularization to induce sparsity.
As to the robustness aspect, our proposed system depends on the robustness of the selected attribute discovery methods, however, further studies on various surveillance datasets are required to fully understand the proposed system robustness.

{\small
\bibliographystyle{ieee}
\bibliography{egbib}
}

\end{document}